\documentclass[11pt,a4paper]{article}
\usepackage[hyperref]{emnlp2020}
\usepackage{times}
\usepackage{latexsym}
\usepackage{tcolorbox}

\usepackage{xpatch}
\usepackage{array}
\usepackage[noend]{algpseudocode}
\usepackage{listing}
\usepackage{graphicx}
\usepackage{subcaption}
\usepackage{longtable}
\usepackage{supertabular}
\usepackage{booktabs}
\usepackage{float}
\usepackage{hyperref}
\usepackage[export]{adjustbox}
\usepackage{breakcites}
\usepackage{color}
\usepackage{marginnote}
\usepackage{fancyvrb}
\usepackage{listings}
\usepackage{xcolor}
\usepackage{enumitem}
\usepackage{anyfontsize}
\usepackage{multirow}
\usepackage{makecell}
\usepackage{amssymb}% http://ctan.org/pkg/amssymb
\usepackage{pifont}% http://ctan.org/pkg/pifont

\usepackage{microtype}

\aclfinalcopy % Uncomment this line for the final submission
\newcommand{\footremember}[2]{%
   \thanks{\xspace\xspace#2}
    \newcounter{#1}
    \setcounter{#1}{\value{footnote}}%
}
\newcommand{\footrecall}[1]{%
    \footnotemark[\value{#1}]%
}

\usepackage{amssymb}
\usepackage{amsmath,amsfonts}
\usepackage{amsopn,bm,mathtools}

\usepackage{nicefrac}       % compact symbols for 1/2, etc.
\newcommand{\vct}[1]{\boldsymbol{#1}} % vector
\newcommand{\mat}[1]{\boldsymbol{#1}} % matrix
\newcommand{\T}{^{\textrm T}} % transpose

\newcommand{\ProbOpr}[1]{\mathbb{#1}}
\newcommand{\expect}[2]{%
\ifthenelse{\equal{#2}{}}{\ProbOpr{E}_{#1}}
{\ifthenelse{\equal{#1}{}}{\ProbOpr{E}\left[#2\right]}{\ProbOpr{E}_{#1}\left[#2\right]}}} % Expectation: syntax: E{1}{2} = E_1[2], E{}{2}=E[2], E{1}{} = E_1
\newcommand{\var}[2]{%
\ifthenelse{\equal{#2}{}}{\ProbOpr{VAR}_{#1}}
{\ifthenelse{\equal{#1}{}}{\ProbOpr{VAR}\left[#2\right]}{\ProbOpr{VAR}_{#1}\left[#2\right]}}} % Expectation: syntax: V{1}{2} = V_1[2], V{}{2}=V[2], V{1}{} = V_1
\DeclareMathOperator{\argmax}{arg\,max}

\newcommand{\vv}{\vct{v}}

\newcommand{\vx}{{\vct{x}}}
\newcommand{\vy}{\vct{y}}

\newcommand{\mX}{\mat{X}}

\newcommand{\sD}{\mathcal{D}}

\newcommand{\sG}{\mathcal{G}}

\newcommand{\vphi}{\vct{\phi}}
\newcommand{\vpsi}{\vct{\psi}}
\newcommand{\vtheta}{\vct{\theta}}

\newcommand{\supp}{the Appendix\xspace}

\newcommand{\ourtitle}{{Learning to Represent Image and Text with Denotation Graph}}
\newcommand{\vl}{\emph{vision\,$+$\,language}\xspace}
\newcommand{\Flickr}{\textsc{flickr30k}\xspace}
\newcommand{\DGFlickr}{\textsc{dg-flickr30k}\xspace}
\newcommand{\COCO}{\textsc{coco}\xspace}
\newcommand{\DGCOCO}{\textsc{dg-coco}\xspace}
\newcommand{\REFCOCO}{\textsc{refcoco+}\xspace}
\newcommand{\MITSTATE}{\textsc{MIT-State}\xspace}

\usepackage{xspace}
\makeatletter
\DeclareRobustCommand\onedot{\futurelet\@let@token\@onedot}
\def\@onedot{\ifx\@let@token.\else.\null\fi\xspace}
\def\eg{\emph{e.g}\onedot} 
\def\ie{\emph{i.e}\onedot}

\def\etal{\emph{et al}\onedot}
\makeatother

\newcommand\mypara[1]{\vspace{1mm}\noindent\textbf{#1}}

\newcommand\paren[1]{\left(#1\right)}
\newcommand\braces[1]{\left\{#1\right\}}

\usepackage{mathtools}

\usepackage[ruled,vlined]{algorithm2e}

\usepackage{listings}

\graphicspath{{../figs/}{../}}

\newcommand{\eat}[1]{}

\begin{document}
\title{\ourtitle}

\author{
  Bowen Zhang \footremember{Google}{Work done while at Google} \footremember{Equal}{Authors Contributed Equally} \\
  University of Southern California \\
  {\tt zhan734@usc.edu} \\\And
  Hexiang Hu \footrecall{Google} \xspace\xspace \footrecall{Equal} \\
  University of Southern California \\
  {\tt hexiangh@usc.edu} \\\AND
  Vihan Jain \\
  Google Research \\
  {\tt vihanjain@google.com} \\\And
  Eugene Ie \\
  Google Research \\
  {\tt eugeneie@google.com} \\\And
  Fei Sha \footremember{leave}{On leave from USC (feisha@usc.edu)} \\
  Google Research \\
  {\tt fsha@google.com}
\\}

\date{}

\maketitle

\begin{abstract}
Learning to fuse vision and language information and representing them is an important research problem with many applications. Recent progresses have leveraged the ideas of pre-training (from language modeling) and attention layers in Transformers to learn representation from datasets containing images aligned with linguistic expressions that describe the images.  In this paper, we propose learning representations from a set of implied, visually grounded expressions between image and  text, automatically mined from those datasets. In particular, we use denotation graphs to represent how specific concepts (such as sentences describing images) can be linked to abstract and generic concepts (such as short phrases) that are also visually grounded. This type of generic-to-specific relations can be discovered using linguistic analysis tools. We propose methods to incorporate such relations into learning representation. We show that state-of-the-art multimodal learning models can be further improved by leveraging automatically harvested structural relations. The representations lead to stronger empirical results on downstream tasks of cross-modal image retrieval, referring expression, and compositional attribute-object recognition. Both our codes and the extracted denotation graphs on the Flickr30K and the COCO datasets are publically available on \url{https://sha-lab.github.io/DG}.
\end{abstract}

\section{Introduction}
\label{sIntro}

There has been an abundant amount of aligned visual and language data such as text passages describing images, narrated videos, subtitles in movies, etc. Thus, learning how to represent visual and language information when they are semantically related has been a very actively studied topic. There are many \vl applications: image retrieval with descriptive sentences or captions~\cite{barnard2001learning,barnard2003matching,hodosh2013framing,young2014image}, image captioning~\cite{chen2015microsoft,xu2015show}, visual question answering~\cite{antol2015vqa}, visual navigation with language instructions~\cite{anderson2018vision},  visual objects localization via short text phrases~\cite{Plummer2015Flickr30kEC}, and others. A  recurring theme is to learn the representation of these two streams of information so that they correspond to each other, highlighting the notion that many language expressions are visually grounded.

A standard approach is to embed the visual and the language information as points in a (joint) visual-semantic embedding space~\cite{frome2013devise,kiros2014UVS,faghri2018vse++}. One can then infer whether the visual information is aligned with the text information by checking how these points are distributed.

How do we embed visual and text information?  Earlier approaches focus on embedding each stream of information independently, using models that are tailored to each modality. For example, for image, the embedding could be the features at the last fully-connected layer from a deep neural network trained for classifying the dominant objects in the image. For text, the embedding could be the last hidden outputs from a recurrent neural network. 

Recent approaches, however, have introduced several innovations~\cite{Lu2019ViLBERTPT,Li2019UnicoderVLAU,chen2019uniter}. The first is to contextualize the embeddings of one modality using information from the other one. This is achieved by using co-attention or cross-attention (in addition to self-attention) in Transformer layers. The second is to leverage the power of pre-training~\cite{radford2019language,Devlin2019BERT}: given a large number of parallel corpora of images and their descriptions, it is beneficial to identify pre-trained embeddings on these data such that they are useful for downstream \vl  tasks.

Despite such progress, there is a missed opportunity of learning stronger representations from those parallel corpora.
As a motivating example, suppose we have two paired examples: one is an image $\vx_1$ corresponding to the text $\vy_1$ of \textsc{two dogs sat in front of porch} and the other is an image $\vx_2$ corresponding to the text  $\vy_2$ of \textsc{two dogs running on the grass}. Existing approaches treat the two pairs independently and compute the embeddings for each pair without acknowledging that both texts share the common phrase $\vy_1 \cap \vy_2= \textsc{two dogs}$ and the images have the same visual categories of two dogs.

We hypothesize that learning the correspondence between the common phrase $\vy_1 \cap \vy_2$ and the set of images $\braces{\vx_1, \vx_2}$, though not explicitly annotated in the training data, is beneficial. Enforcing the alignment due to this \emph{additionally constructed} pair introduces a form of structural constraint: the embeddings of $\vx_1$ and $\vx_2$ have to convey similar visual information that is congruent to the similar text information in the embeddings of $\vy_1$ and $\vy_2$.

In this paper, we validate this hypothesis and show that extracting additional and implied correspondences between the texts and the visual information, then using them for learning leads to better representation, which results in a stronger performance in downstream tasks. The additional alignment information forms a graph where the edges indicate how visually grounded concepts can be instantiated at both abstract levels (such as \textsc{two dogs}) and specific levels (such as \textsc{two dogs sat in front of the porch}). These edges and the nodes that represent the concepts at different abstraction levels form a graph, known as denotation graph, previously studied in the NLP community~\cite{young2014image,lai2017learning,Plummer2015Flickr30kEC} for grounding language expressions visually.

Our contributions are to propose creating visually-grounded denotation graphs to facilitate representation learning. Concretely, we apply the technique originally developed for the \Flickr dataset~\cite{young2014image} also to \COCO dataset~\cite{lin2014microsoft} to obtain denotation graphs that are grounded in each domain respectively (\S~\ref{sDgraph}). We then show how the denotation graphs can be used to augment training samples for aligning text and image (\S~\ref{sMethod}). Finally, we show empirically that the representation learned with denotation graphs leads to stronger performance in downstream tasks (\S~\ref{sExp}).
\begin{figure*}[t]
    \centering
    \begin{tabular}{c@{\quad\quad\quad\quad}c}
        \includegraphics[height=4cm]{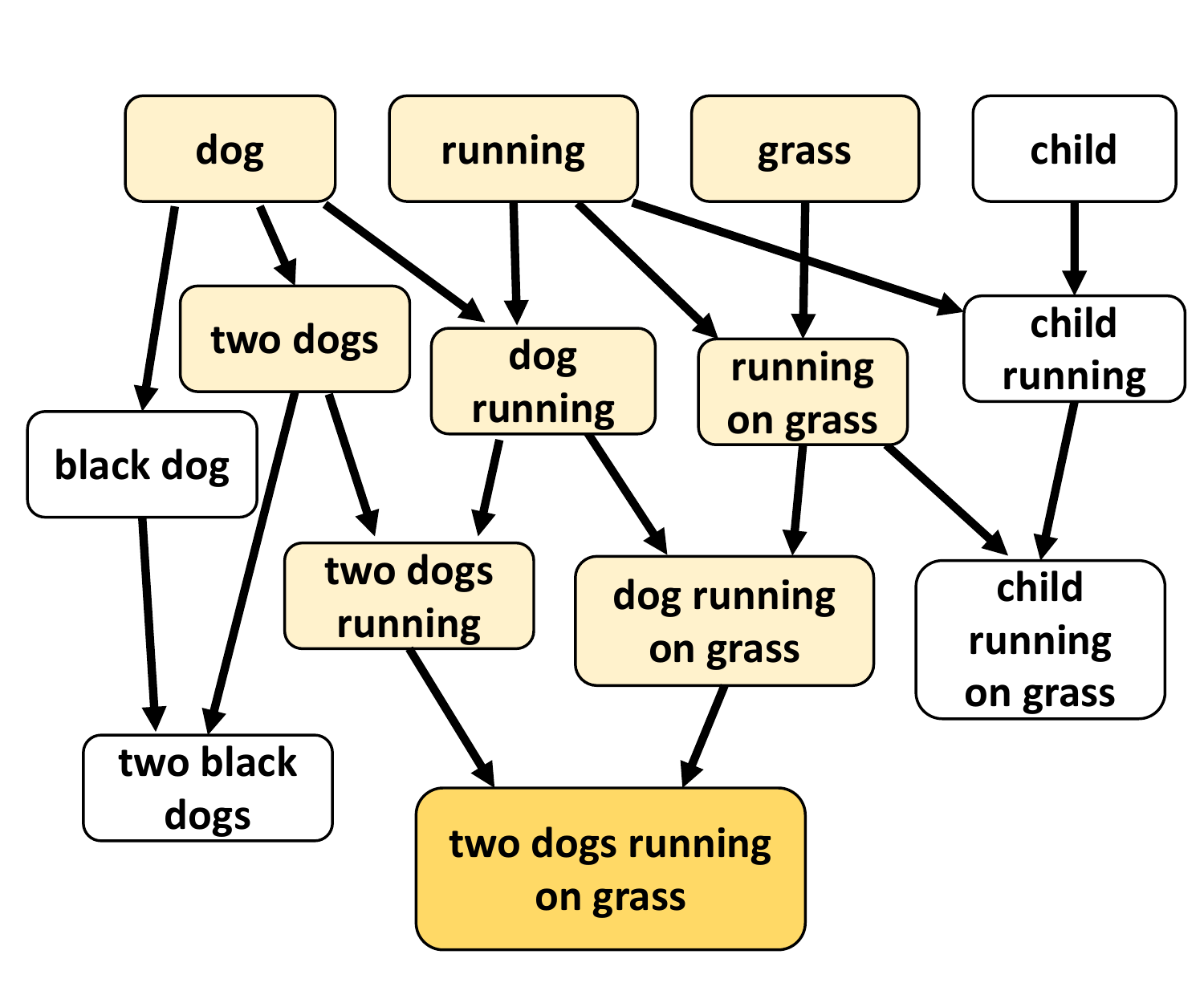} & \includegraphics[height=4cm]{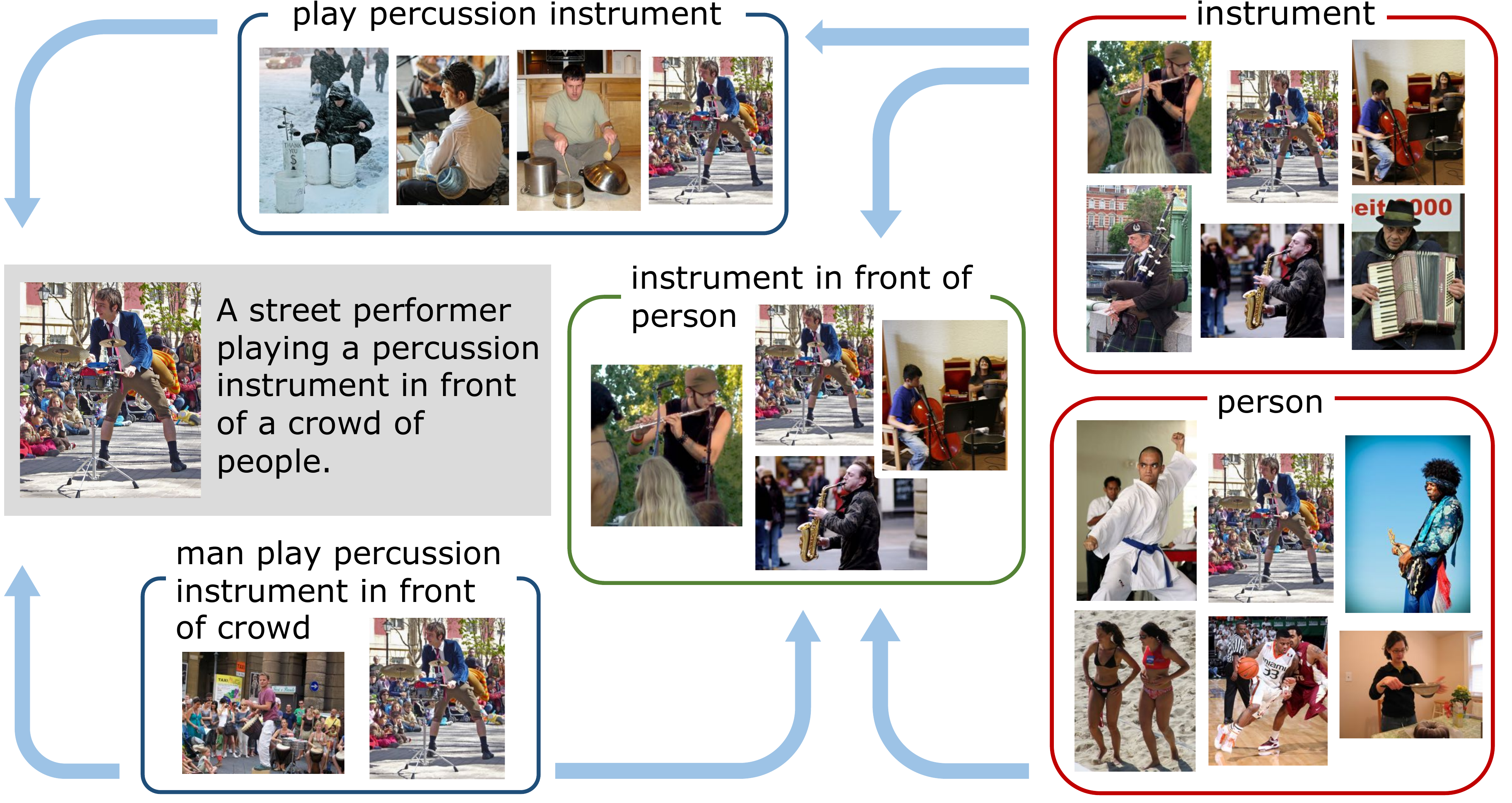}
    \end{tabular}
    \caption{(Left) A schematic example of denotation graph showing the hierarchical organization of linguistic expression (adapted from \url{https://shannon.cs.illinois.edu/DenotationGraph/}) (Right) A random-subgraph from the denotation graph extracted from the \Flickr dataset, with images attached to concepts at different levels of hierarchy.
    }
    \label{fig:DG}
\end{figure*}
% !TEX root = main_emnlp2020.tex

\section{Related Work}
\label{sec:related}

\par\noindent\mypara{Learning representation for image and text} Single-stream methods learn each modality separately and align them together with a simple fusion model, often an inner product between the two representations. Frome~\etal~\cite{frome2013devise} learns the joint embedding space for images and labels and use the learned embeddings for zero-shot learning. Kiros~\etal~\cite{kiros2014UVS} uses bi-directional LSTMs to encode sentences and then maps images and sentences into a joint embedding space for cross-modal retrieval and multi-modal language models.
Li~\etal~\cite{Li2019VSRN} designs a high-level visual reasoning module to contextualize image entity features and obtain a more powerful image representation. Vendrov~\etal~\cite{vendrov2015order} improves image retrieval performance by exploiting the hypernym relations among words. There is a large body of work that has been focusing on improving the visual or text embedding functions~\cite{Socher2014GroundedCS,eisenschtat20172waynet,nam2017DAN,huang2018SCO,gu2018GXN}.

Another line of work,  referred to as cross-stream methods infer fine-grained alignments between local patterns of visual (\ie, local regions) and linguistic inputs (\ie, words) between a pair of image and text, then use them to derive the similarity between the image and the text. SCAN~\cite{lee2018scan} uses cross-modal attention mechanism~\cite{xu2015show} to discover such latent alignments. Inspired by the success of BERT~\cite{Devlin2019BERT}, recent efforts have conducted visual-linguistic pre-training on large-scale datasets~\cite{sharma2018conceptual}, using a powerful sequence model such as deep Transformers~\cite{Lu2019ViLBERTPT,Li2019UnicoderVLAU,chen2019uniter,su2019vl,li2019visualbert}.  The pre-training strategies of these methods typically involve many self-supervised learning tasks, including the image-text matching~\cite{Lu2019ViLBERTPT}, masked language modeling~\cite{Devlin2019BERT,Lu2019ViLBERTPT} and masked region modeling~\cite{chen2019uniter}.

In contrast to those work, we focus on exploiting additional correspondences between image and text that are not explicitly given in the many image and text datasets. By analyzing the linguistic structures of the texts in those datasets, we are able to discover more correspondences that can be used for learning representation. We show the learned representation is more powerful in downstream tasks.

\par\noindent\mypara{\emph{Vision\,$+$\,Language} Tasks} There has been a large collection of  tasks combining  vision and  language, including \emph{image captioning}~\cite{chen2015mind,fang2015captions,hodosh2013framing,karpathy2015deep,kulkarni2013babytalk}, \emph{visual QA}~\cite{antol2015vqa}, \emph{text-based image verification}~\cite{suhr2017corpus,suhr2018corpus,hu2019evaluating}, \emph{visual commonsense reasonin}~\cite{zellers2019recognition}, and so on. In the context of this paper, we focus on studying \emph{cross-modality retrieval}~\cite{barnard2003matching,barnard2001learning,gong2014multi,hodosh2013framing,young2014image,zhang2018cross}, as well as transfer learning on downstream tasks, including \emph{compositional attribute-object recognition}~\cite{isola2015discovering,misra2017red} and \emph{referring expressions}~\cite{dale1995computational,kazemzadeh2014referitgame,kong2014you,mitchell2012midge}. Please refer to \S~\ref{sExp} for explanation of these tasks.  

% !TEX root = main_emnlp2020.tex
\section{Denotation Graph (DG)}
\label{sDgraph}

Visually grounded text expressions denote the images (or videos) they describe. When examined together, these expressions reveal structural relations that do not exhibit when each expression is studied in isolation. In particular, through linguistic analysis, these expressions can be grouped and partially ordered and thus form a relation graph, representing how (visually grounded) concepts are shared among different expressions and how different concepts are related.  This insight was explored by \citet{young2014image} and the resulting graph is referred to as a denotation graph, schematically shown in the top part of Fig.~\ref{fig:DG}. In this work, we focus on constructing denotation graphs from the \Flickr and the \COCO datasets, where the text expressions are sentences describing images.

Formally, a denotation graph $\sG$ is a polytree where a node $v_i$ in the graph corresponds to  a pair of a linguistic expression $\vy_i$ and a set of images $\mX_i =\braces{\vx_1, \vx_2, \cdots, \vx_{n_i}}$.  A directed edge $e_{ij}$ from a node $v_i$ to its child $v_j$ represents a subsumption relation between $\vy_i$ and $\vy_j$. Semantically, $\vy_i$ is more abstract (generic) than $\vy_j$, and the tokens in $\vy_i$ can be a subset of $\vy_j$'s. For example, \textsc{two dogs} describes all the images which  \textsc{two dogs are running} describes, though less specifically. Note that the subsumption relation is defined on the semantics of these expressions. Thus, the tokens do not have to be exactly matched on their surface forms. For instance, \textsc{in front of person} or \textsc{in front of crowd} are also generic concepts that subsume \textsc{in front of a crowd of people}, see the right-hand side of Fig.~\ref{fig:DG} for another example.

More formally, the set of images that correspond to $\vv_i$ is the union of all the images corresponding to $\vv_i$'s children $\textrm{ch}(v_i)$: $\mX_i = \bigcup_{v_j \in \textrm{ch}(v_i)} \mX_j$. We also use $\textrm{pa}(v_j)$ to denote the set of  $v_j$'s parents.

Denotation graphs  (DG) can be seen as a hierarchical organization of semantic knowledge among concepts and their visual groundings. In this sense, they generalize the tree-structured object hierarchies that have been often used in computer vision. The nodes in the DG are composite phrases that are semantically richer than object names and the relationship among them is also richer.

\begin{table}[t]
    \small
    \centering
    \caption{Key statistics of the two DGs: averaged over the all nodes in the graph, internal nodes and leaf nodes (formated as all/internal/leaf)}
    \label{tDGs}
    \resizebox{\linewidth}{!}{
        \begin{tabular}{@{\;} l@{\;\;} c@{\;\;} c@{\;\;} c@{\;\;} c@{\;}}
            \toprule
            Dataset & \DGFlickr & \DGCOCO \\
            \midrule
            \# of edges & 1.94$M$ & 4.57$M$ \\
            \# of nodes & 597$K$/452$K$/145$K$ & 1.41$M$/841$K$/566$K$ \\
            \# of tokens/node & 6.78/4.45/14.04 & 5.88/4.07/8.58 \\
            \# of images/node & 4.46/5.57/1.00 & 5.06/7.79/1.00\\
            \bottomrule
        \end{tabular}
    }
\end{table}

\mypara{Constructing DG} We used the publicly available tool\footnote{Available online at \url{https://github.com/aylai/DenotationGraph}}, following Young~\etal~\cite{young2014image}. For details, please refer to \supp and the reference therein. Once the graph is constructed, we attach the images to the proper nodes by set-union images of each node's children, starting from the sentence-level node.

\mypara{\DGFlickr and \DGCOCO}\footnote{Both DGs are made publically available at \url{https://sha-lab.github.io/DG/}} We regenerate a DG on the \Flickr dataset\footnote{The original DG, while publicly available at \url{https://shannon.cs.illinois.edu/DenotationGraph/} contains 1.75 million nodes which are significantly less than ours, due to the difference in the version of the NLP toolkit.}~\cite{young2014image} and construct a new DG on the COCO~\cite{lin2014microsoft} dataset. The two datasets come from different visual and text domains where the former contains more iconic social media photos and the latter focuses on photos with complex scenes and has more objects. Figure~\ref{fig:DG} shows a random sub-graph of \DGFlickr.

Table~\ref{tDGs} lists the key statistics of the two DGs. We note that in both graphs, a large number of internal nodes (more abstract concepts or phrases) are introduced. For such concepts, the linguistic expressions are much shorter and the number of images they correspond to is also larger.
% !TEX root = main_emnlp2020.tex
\section{Learning with Denotation Graphs}
\label{sMethod}

The denotation graphs, as described in the previous section, provide rich structures for learning representations of text and image. In what follows, we describe three learning objectives, starting from the most obvious one that matches images and their descriptions (\S~\ref{subsec:learning_itm}), followed by learning to discriminate between general and specialized concepts (\S~\ref{subsec:abstract}) and learning to predict concept relatedness~(\S~\ref{subsec:objective_relation}). We perform ablation studies of those objectives in \S~\ref{sAblation}.

\subsection{Matching Texts with Images}
\label{subsec:learning_itm}

We suppose the image $\vx$ and the text $\vy$ are represented by (a set of) vectors $\vphi(\vx)$ and $\vpsi(\vy)$ respectively. A common choice for $\vphi(\cdot)$ is the last layer of a convolutional neural network~\cite{He2015ResNet,Xie2016ResNeXt} and for $\vpsi(\cdot)$ the contextualized word embeddings from a Transformer network~\cite{Vaswani2017Transformer}. The embedding of the \emph{multimodal} pair is a vector-valued function over $\vphi(\vx)$ and $\vpsi(\vy)$:
\begin{equation}
\vv(\vx, \vy) = f ( \vphi(\vx), \vpsi(\vy))
\end{equation}
There are many choices of $f(\cdot, \cdot)$. The simplest one is to concatenate the two arguments. We can also use the element-wise product between the two if they have the same embedding dimension~\cite{kiros2014UVS}, or complex mappings parameterized by layers of attention networks and convolutions~\cite{Lu2019ViLBERTPT,chen2019uniter} -- we experimented some of them in our empirical studies.

\subsubsection{Matching Model}
We use the following probabilistic model to characterize the joint distribution
\begin{equation}
p(\vx, \vy) \propto \exp ( \vtheta\T \vv(\vx, \vy))
\end{equation}
where the exponent $s(\vx, \vy) = \vtheta\T \vv$ is referred as the matching score. To estimate $\vtheta$, we use the maximum likelihood estimation
\begin{equation}
\vtheta^* = \argmax  \sum\nolimits_{v_i} \sum\nolimits_k \log p(\vx_{ik}, \vy_i)
\end{equation}
where $\vx_{ik}$ is the $k$th element in the set $\mX_i$. However, this probability is intractable to compute as it requires us to get all possible pairs of $(\vx, \vy)$. To approximate, we use negative sampling.

\subsubsection{Negative Sampling} For each (randomly selected) positive sample $(\vx_{ij}, \vy_i)$, we explore 4 types of negative examples and assemble them as a negative sample set $\sD_{ik}^-$:

\noindent
\emph{Visually mismatched pair}\quad  We randomly sample an image $\vx^- \notin \mX_i$ to pair with $\vy_i$, \ie, $\paren{\vx^-, \vy_i}$. Note that we automatically exclude the images from $v_i$'s children.\\[0.5em]
\emph{Semantically mismatched pair}\quad We randomly sample a text $\vy_j \ne \vy_i$ to form the pair $\paren{\vx_{ik}, \vy_j}$. Note that we constrain $\vy_j$ not to include concepts that could be more abstract than $\vy_i$ as the more abstract can certainly be used to describe the specific images $\vx_{ik}$.\\[0.5em]
\emph{Semantically hard pair}\quad We randomly sample a text $\vy_j$ that corresponds to an image $\vx_j$ that is visually similar to $\vx_{ik}$ to form $\paren{\vx_{ik}, \vy_j}$. See \cite{Lu2019ViLBERTPT} for details.\\[0.5em]
\emph{DG Hard Negatives}\quad We randomly sample a \emph{sibling} (but not cousin) node $v_j$ to $v_i$ such that  $\vx_{ik} \notin \mX_j$ to form $\paren{\vx_{ik}, \vy_j}$

Note that the last 3 pairs have increasing degrees of semantic confusability. In particular, the 4\emph{th} type of negative sampling is only possible with the help of a denotation graph. In that type of negative samples, $\vy_j$ is semantically very close to $\vy_i$ (from the construction) yet they denote different images. The ``semantically hard pair'', on the other end, is not as hard as the last type as $\vy_i$ and $\vy_j$ could be very different despite high visual similarity.

With the negative samples, we estimate $\vtheta$ as the minimizer of the following negative log-likelihood
\begin{equation}
\ell_\textsc{match}= - \sum_{v_i} \sum_k \log \frac{e^{s(\vx_{ik}, \vy_i)}}{\sum_{(\hat{\vx}, \hat{\vy}) \sim \sD_i}e^{s(\hat{\vx}, \hat{\vy})}}
\end{equation}
where $\sD_i  = \sD^-_{ik} \cup \braces{ \paren{\vx_{ik}, \vy_i}}$ contains both the positive and negative examples.
\subsection{Learning to Be More Specific}
\label{subsec:abstract}

The hierarchy in the denotation graph introduces an opportunity for learning image and text representations that are sensitive to fine-grained distinctions.
Concretely, consider a parent node $\vv_i$ with an edge to the child node $\vv_j$. While the description $\vy_j$ matches any images in its children nodes, the parent node's description $\vy_i$ on a higher level is more abstract. For example, the concepts \textsc{instrument} and \textsc{play percussion instrument} in Fig~\ref{fig:DG} is a pair of examples showing the latter more accurately describes the image(s) at the lower-level.

To incorporate this modeling notion, we introduce 
\begin{align}
    \ell_{\textsc{spec}} = \sum_{e_{ij}}\sum_k [ s(\vx_{jk}, \vy_i) - s(\vx_{jk}, \vy_j) ]_+
    \label{obj:abstract}
\end{align}
as a specificity loss, where $[h]_+ = \text{max}(0, h)$ denotes the hinge loss. The loss is to be minimized such that the matching score for the less specific description  $\vy_i$ is smaller than that for the more specific description $\vy_j$.

\subsection{Learning to Predict Structures}
\label{subsec:objective_relation}

Given the graph structure of the denotation graph, we can also improve the accuracy of image and text representation by modeling high-order relationships. Specifically, for a pair of nodes $v_i$ and $v_j$, we want to predict whether there is an edge from $v_i$ to $v_j$, based on each node's corresponding embedding of a pair of image and text. Concretely, this is achieved by minimizing the following negated likelihood
\begin{align}
\ell_{\textsc{edge}} = -\sum\nolimits_{e_{ij}}&  \sum\nolimits_{k, k'} \log p( e_{ij}=1| \nonumber \\
&   \vv(\vx_{ik}, \vy_i), \vv(\vx_{jk'}, \vy_j))
\end{align}
We use a multi-layer perceptron with a binary output to parameterize the log-probability.

\subsection{The Final Learning Objective}
\label{subsec:objective_overall}

We combine the above loss functions as the final learning objective for learning on the DG
\begin{align}
        \ell_{\textsc{dg}} = \ell_{\textsc{match}} + \lambda_1 \cdot \ell_{\textsc{spec}} + \lambda_2 \cdot \ell_{\textsc{edge}}
        \label{eLossDG}
\end{align}
where $\lambda_1$, $\lambda_2$ are the hyper-parameters that trade-off different losses. Setting them to 1.0 seems to work well. The performance under different $\lambda_1$ and $\lambda_2$ are reported in Table \ref{tab:vilbert_reduced} and Table \ref{tab:vilbert}. We study how each component could affect the learning of representation in \S~\ref{sAblation}.

% !TEX root = main_emnlp2020.tex
\section{Experiments}
\label{sExp}

We examine the effectiveness of using denotation graphs to learn image and text representations. We first describe the experimental setup and key implementation details (\S~\ref{subsec:setup}). We then describe key image-text matching results in \S~\ref{exp:results}, followed by studies about the transfer capability of our learned representation (\S~\ref{exp:transfer}). Next, we present ablation studies over different components of our model (\S~\ref{sAblation}). Finally, we validate how well abstract concepts can be used to retrieve images, using our model (\S~\ref{sGeneric}).

\subsection{Experimental Setup}
\label{subsec:setup}

We list major details in the following to provide context, with the full details documented in \supp for reproducibility.

\mypara{Embeddings and Matching Models} Our aim is to show denotation graphs  improve state-of-the-art methods. To this end, we experiment  with two recently proposed state-of-the-art approaches and their variants for learning from multi-modal data: ViLBERT~\cite{Lu2019ViLBERTPT} and UNITER~\cite{chen2019uniter}. The architecture diagrams and the implementation details are in \supp, with key elements summarized in the following.

Both the approaches start with an image encoder, which obtains a set of embeddings of image patches, and a text encoder which obtains a sequence of word (or word-piece) embeddings. For ViLBERT, text tokens are processed with  Transformer layers and fused with the image information with 6 layers of co-attention Transformers. The output of each stream is then element-wise multiplied to give the fused embedding of both streams. For UNITER, both streams are fed into 12 Transformer layers with cross-modal attention. A special token \textsc{cls} is used, and its embedding is regarded as the fused embedding of both streams.

For ablation studies, we use a smaller ViLBERT for rapid experimentation: ViLBERT (Reduced) where there are 3 Transformer layers and 2 co-attention Transformers for the text stream, and 1 Transformer layer for the image stream.

\mypara{Constructing Denotation Graphs}  As described in \S\ref{sDgraph}, we construct denotation graphs \DGFlickr and \DGCOCO from the \textbf{\Flickr}~\cite{young2014image} and the \textbf{\COCO}~\cite{lin2014microsoft} datasets. \Flickr was originally developed for the tasks of image-based and text-based retrieval. It contains 29,000 images for training, 1,000 images for validation, and 1,000 images for testing. \COCO is a significantly larger dataset, developed for the image captioning task. It contains 565,515 sentences with 113,103 images. We evaluate on both the 1,000 images testing split and the 5,000 images testing split (in \supp), following the setup in~\citep{karpathy2015deep}.  Key characteristics for the two DGs are reported in Table~\ref{tDGs}.

\mypara{Evaluation Tasks} We evaluate the learned representations on three common \vl tasks. In text-based image retrieval, we evaluate two settings: the text is either a sentence or a phrase from the test corpus. In the former setting, the sentence is a leaf node on the denotation graph, and in the latter case, the phrase is an inner node on the denotation graph, representing more general concepts. We evaluate the \Flickr and the \COCO datasets, respectively. The main evaluation metrics we use are precisions at recall \textsc{r@m} where $\textsc{m}=1, 5$ or $10$ and \textsc{rsum} which is the sum of the 3 precisions~\cite{Wu2019UniVSE}. Conversely, we also evaluate using the task of image-based text retrieval to retrieve the right descriptive text for an image.

In addition to the above cross-modal retrieval, we also consider two downstream evaluation tasks, \ie, \textbf{Referring Expression} and \textbf{Compositional Attribute-Object Recognition}. (1) Referring Expression is a task where the goal is to localize the corresponding object in the image given an expression~\cite{kazemzadeh2014referitgame}. We evaluate on the dataset \textbf{\REFCOCO}, which contains 141,564 expressions with 19,992 images. We follow the previously established protocol to evaluate on the validation split, the TestA split, and the TestB split. We are primarily interested in zero-shot/few-shot learning performance. (2) Compositional Attribute-Object Recognition is a task that requires a model to learn from images of {SEEN} (attribute, object) label pairs, such that it can generalize to recognize images of {UNSEEN} (attribute, object) label pairs. We evaluate this task on the \textbf{\MITSTATE} dataset~\cite{isola2015discovering}, following the protocol by~\citet{misra2017red}. The training split contains 34,562 images from 1,262 SEEN labels, and the test split contains 19,191 images from 700 UNSEEN labels. We report the Top-1, 2, 3 accuracies on the UNSEEN test set as evaluation metrics.

\begin{table}[t]
    \centering
    \small
    \caption{Text-based Image Retrieval (Higher is better)}
    \begin{tabular}{@{\;}lcccc@{\;}}
        %\addlinespace
        \toprule
        Method~ &\textsc{r@1} & \textsc{r@5} & \textsc{r@10} & \textsc{rsum} \\
        \midrule
        \multicolumn{5}{@{\;}c}{\bf \Flickr} \\
        ViLBERT & 59.1 & 85.7 & 92.0 & 236.7  \\
       	ViLBERT + DG & 63.8 & 87.3 & 92.2 & 243.3 \\
       	UNITER  & 62.9 & 87.2 & 92.7 & 242.8 \\
        UNITER + DG & 66.4 & 88.2 & 92.2 & 246.8 \\
        \midrule
        \multicolumn{5}{@{\;}c}{\bf \COCO 1K Test Split}\\
        ViLBERT & 62.3 & 89.5 & 95.0 & 246.8 \\
        ViLBERT + DG & 65.9 & 91.4 & 95.5 & 252.7 \\
        UNITER  &  60.7 & 88.0 & 93.8 & 242.5 \\
        UNITER + DG &  62.7 & 88.8 & 94.4 & 245.9 \\
        \midrule
        \multicolumn{5}{@{\;}c}{\bf \COCO 5K Test Split}\\
        ViLBERT & 38.6 & 68.2 & 79.0 & 185.7 \\
        ViLBERT + DG & 41.8 & 71.5 & 81.5 & 194.8 \\
        UNITER & 37.8 & 67.3 & 78.0 & 183.1 \\
        UNITER + DG &  39.1 & 68.0 & 78.3 & 185.4 \\
        \bottomrule
    \end{tabular}
    \vspace{-0.5em}
    \label{tText2Image}
\end{table}

\mypara{Training Details} Both ViLBERT and UNITER models are pre-trained on the Conceptual Caption dataset~\cite{sharma2018conceptual} and the pre-trained models are released publicly\footnote{The UNITER\cite{chen2019uniter} model performs an additional online hard-negative mining (which we did not) during the training of image-text matching to improve their results. This is computationally very costly.}. On the \DGFlickr, ViLBERT and UNITER are trained with a minibatch size of 64 and ViLBERT is trained for 17 epochs and UNITER for 15 epochs, with a learning rate of $0.00004$. On the \DGCOCO, ViLBERT is trained for 17 epochs and UNITER for 15 epochs with a minibatch size of 64 and a learning rate of $0.00004$. The hyperparameters in Eq.~(\ref{eLossDG}) are set to 1.0, unless specified (see \supp).

\subsection{Main Results}
\label{exp:results}

Table~\ref{tText2Image} and Table~\ref{tImage2Text} report the performances on cross-modal retrieval. On both datasets, models trained with denotation graphs considerably outperform the corresponding ones which are not.

For the image-based text retrieval task, ViLBERT and UNITER on \Flickr suffers a small drop in \textsc{r@10} when DG is used. On the same task, UNITER on \COCO 5K Test Split decreases more when DG is used. However, note that on both splits of \COCO, ViLBERT is a noticeably stronger model, and using DG improves its performance.

\subsection{Zero/Few-Shot and Transfer Learning}
\label{exp:transfer}

\begin{table}[t]
    \centering
    \small
    \caption{Image-based Text Retrieval (Higher is better)}
    \begin{tabular}{@{\;}l@{\quad}cccc @{\;}}
        \toprule
        Method &\textsc{r@1} & \textsc{r@5} & \textsc{r@10} & \textsc{rsum} \\
        \midrule
        \multicolumn{5}{@{\;}c}{\bf \Flickr} \\
    	ViLBERT     &  76.8 & 93.7 & 97.6 & 268.1\\
       	ViLBERT + DG & 77.0 & 93.0 & 95.0 & 265.0 \\
        UNITER  &  78.3 & 93.3 & 96.5 & 268.1 \\
        UNITER + DG &  78.2 & 93.0 & 95.9 & 267.1 \\
       	\midrule
        \multicolumn{5}{@{\;}c}{\bf \COCO 1K Test Split} \\
        ViLBERT      &  77.0 & 94.1 & 97.2 & 268.3 \\
        ViLBERT + DG & 79.0 & 96.2 & 98.6 & 273.8\\
        UNITER       & 74.4 & 93.9 & 97.1 & 265.4 \\
        UNITER + DG  & 77.7 & 95.0 & 97.5 & 270.2 \\
        \midrule
        \multicolumn{5}{@{\;}c}{\bf \COCO 5K Test Split} \\
        ViLBERT      & 53.5 & 79.7 & 87.9 & 221.1\\
        ViLBERT + DG & 57.5 & 84.0 & 90.1 & 232.2 \\
        UNITER       & 52.8 & 79.7 & 87.8 & 220.3 \\
        UNITER + DG  & 51.4 & 78.7 & 87.0 & 217.1 \\
        \bottomrule
    \end{tabular}
    \label{tImage2Text}
\end{table}

\begin{table}[t]
    \centering
    \small
    \caption{Image Retrieval via Text (Transfer Learning)}
    \begin{tabular}{ l cc  cc }
        \toprule
        \textsc{source} & \multicolumn{2}{c}{\textsc{flickr}$\rightarrow$\COCO} & \multicolumn{2}{c}{\COCO$\rightarrow$\textsc{flickr}} \\
        \quad$\rightarrow$\textsc{target} & \textsc{r@1} & \textsc{rsum} & \textsc{r@1}  & \textsc{rsum} \\
        \midrule
        ViLBERT &  43.5 & 199.5 & 49.0 &  209.0 \\
        \quad + \textsc{source} DG & 44.9  & 200.5 & 52.8 &  218.2 \\
        \bottomrule
    \end{tabular}
    \label{tTransferEval}
\end{table}

\mypara{Transfer across Datasets} Table~\ref{tTransferEval} illustrates that the learned representations assisted by the DG  have better transferability when applied to another dataset (\textsc{target domain}) that is different from the  \textsc{source domain} dataset which the DG is based on. Note that the representations are \emph{not} fine-tuned on the \textsc{target domain}. The improvement on the direction \COCO$\rightarrow$\Flickr is stronger than the reverse one, presumably because the \COCO dataset is bigger than \Flickr. (\textsc{r@5} and \textsc{r@10} are reported in \supp.)

\begin{table*}[t]
    \centering
    \small
	\tabcolsep 5pt
    \caption{Zero/Few-shot Learning for Referring Expression (Reported in \textsc{r@1} on validation, TestA and TestB data)}
    \begin{tabular}{l ccc ccc ccc ccc ccc}
        \toprule
        Setting $\rightarrow$ & \multicolumn{3}{c}{0\% (Zero-shot)} & \multicolumn{3}{c}{25\%} & \multicolumn{3}{c}{50\%} & \multicolumn{3}{c}{100\%}\\

        Method & Val & TestA & TestB & Val & TestA & TestB & Val & TestA & TestB & Val & TestA & TestB\\ \midrule
        ViLBERT  & 35.7  & 41.8 & 29.5 & 67.2 & 74.0 & 57.1  &  68.8 & 75.6 & 59.4 & 71.0 & 76.8 & 61.1 \\
        ViLBERT + \DGCOCO & 36.1  & 43.3 & 29.6 & 67.4 & 74.5 & 57.3  & 69.3 & 76.6 & 59.3 & 71.0 & 77.0 & 60.8 \\
        \bottomrule
    \end{tabular}
    \label{tZeroFewShot}
    \vspace{-0.5em}
\end{table*}

\mypara{Zero/Few-shot Learning for Referring Expression} We evaluate our model on the task of referring expression, a supervised learning task, in the setting of zero/few-shot transfer learning. In zero-shot learning,  we didn't fine-tune the model on the referring expression dataset (\ie \REFCOCO).
Instead, we performed a ``counterfactual'' inference, where we measure the drop in the compatibility score (between a text describing the referring object and the image of all candidate regions) as we removed individual candidates results. The region that causes the biggest drop of compatibility score is selected.
As a result, the selected region is most likely to correspond to the description. In the setting of few-shot learning, we fine-tune our \COCO-pre-trained model on the task of referring expression in an end-to-end fashion on the referring expression dataset (\ie \REFCOCO).

The results in Table~\ref{tZeroFewShot} suggest that when the amount of labeled data is limited, training with DG performs better than training without. When the amount of data is sufficient for end-to-end training, the advantage of training with DG diminishes.

\begin{table}[t]
    \centering
    \small
    \tabcolsep 5pt
    \caption{
    	Image Recognition on UNSEEN Attribute-Object Pairs on the \MITSTATE Dataset
    }
     \begin{tabular}{@{\;\;}l@{\quad}c c c@{\;\;}}
        \toprule
        Method & Top-1 & Top-2 & Top-3 \\
        \midrule
        VisProd~\citep{misra2017red} & 13.6 & 16.1 & 20.6 \\
        RedWine~\citep{misra2017red} & 12.1 & 21.2 & 27.6 \\
        SymNet~\citep{li2020symmetry} & 19.9 & 28.2 & 33.8 \\
        \midrule
        \multicolumn{3}{@{}l}{\bf ViLBERT pre-trained on} \\
        N/A  & 16.2 & 26.3 & 33.3  \\
        \COCO & 17.9 & 28.8 & 36.2 \\
        \DGCOCO & 19.4 & 30.4 & 37.6 \\
        \bottomrule
    \end{tabular}
    \label{tab:att_obj}
\end{table}

\mypara{Compositional Attribute-Object Recognition} We evaluate our model for supervised compositional attribute-object recognition~\cite{misra2017red}, and report results on recognizing UNSEEN attribute-object labels on the \MITSTATE test  data~\cite{isola2015discovering}. Specifically, we treat the text of image labels (\ie, attribute-object pairs as compound phrases) as the sentences to fine-tune the ViLBERT models, using the $\ell_{\textsc{match}}$ objective. Table~\ref{tab:att_obj} reports the results (in top-K accuracies) of both prior methods and variants of ViLBERT, which are trained from scratch (N/A), pre-trained on \COCO and \DGCOCO, respectively. ViLBERT models pre-trained with parallel pairs of images and texts (\ie, \COCO and \DGCOCO) improve significantly over the baseline that is trained on the \MITSTATE from scratch. The model pre-trained with \DGCOCO achives the best results among ViLBERT variants. It performs on par with the previous state-of-the-art method in top-1 accuracy and outperforms them in top-2 and top-3 accuracies.

\begin{table}[t]
    \centering
    \small
    \tabcolsep 10pt
    \caption{
    	Ablation Studies of  Learning from DG
    }
     \begin{tabular}{@{\;\;}l@{\quad}c c@{\;\;}}
        \toprule
        ViLBERT variants $\rightarrow$ & Reduced & Full \\
        \midrule
        {\bf w/o DG} & 215.4 & 236.7 \\
        \midrule
        {\bf w/ DG} & \\
        $+\ \ell_{\textsc{match}}$  & 221.5 & 236.5 \\
          \ \ $-$ \textsc{dg hard negatives}\\
        $+\ \ell_{\textsc{match}}$ & 228.4 & 241.7\\
        $+\ \ell_{\textsc{match}} + \ell_{\textsc{spec}}$ & 228.8 & 242.6\\
        $+\ \ell_{\textsc{match}} + \ell_{\textsc{spec}} + \ell_{\textsc{edge}}$ & 231.2 & 243.3 \\
        \bottomrule
    \end{tabular}
    \label{tab:flickr_abs}
\end{table}

\subsection{Ablation Studies}
\label{sAblation}

The rich structures encoded in the DGs give rise to several components that can be incorporated into learning representations. We study whether they are beneficial to the performances on the downstream task of text-based image retrieval. In the notions of \S\ref{sMethod}, those components are: (1) remove ``\textsc{dg hard negatives}'' from the $\ell_{\textsc{match}}$ loss and only use the other 3 types of negative samples (\S~\ref{subsec:learning_itm}); (2) align images with more specific text descriptions (\S~\ref{subsec:abstract}); (3) predict the existences of edges between pairs of nodes (\S~\ref{subsec:objective_relation}).

Table~\ref{tab:flickr_abs} shows the results from the ablation studies. We report results on two versions of ViLBERT: In ViLBERT (reduced), the number of parameters in the model is significantly reduced by making the model less deep, and thus faster for development. Instead of being pre-trained, they are trained on the \Flickr dataset directly for 15 epochs with a minibatch size of 96 and a learning rate of $4e^{-5}$. In ViLBERT (Full), we use the aforementioned settings. We report \textsc{rsum} on the \Flickr dataset for the task of text-based image retrieval.

All models with DG perform better than the models without DG. Secondly, the components of \textsc{dg hard negatives}, $\ell_{\textsc{spec}}$, and $\ell_\textsc{edge}$ contribute positively and their gains are cumulative.

\subsection{Image Retrieval from Abstract Concepts}
\label{sGeneric}

\begin{figure}[t]
            \centering
            \includegraphics[width=0.4\textwidth]{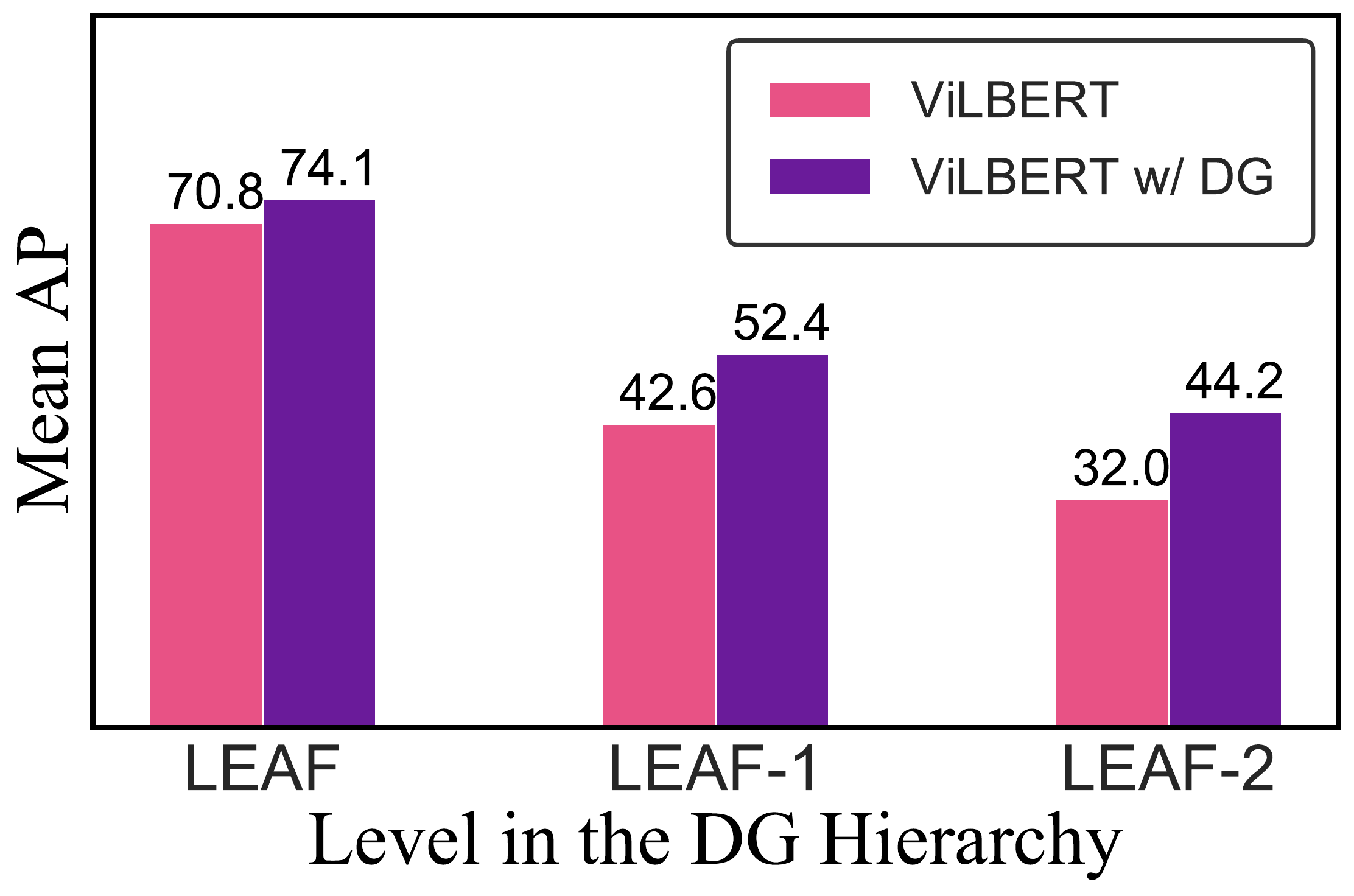}
       \caption{\small
        Image Retrieval using Mid-level Linguistic Expression on \Flickr Denotation Graph. The results are reported in Mean Average Precision (Mean AP).
    }
    \label{fig:phrase}
\end{figure}

The leaf nodes in a DG correspond to complete sentences describing images. The inner nodes are shorter phrases that describe more abstract concepts and correspond to a broader set of images, refer to Table~\ref{fig:phrase} for some key statistics in this aspect.

Fig.~\ref{fig:phrase} contrasts how well abstract concepts can be used to retrieve images. The concepts are the language expressions corresponding to the leaf nodes, the nodes that are one level above (\textsc{leaf}-1), or two levels above (\textsc{leaf}-2) the leaf nodes from the \DGFlickr. Since abstract concepts tend to correspond to multiple images, we use mean averaged precision (mAP) to measure the retrieval results. ViLBERT+DG outperforms ViLBERT significantly. The improvement is also stronger when the concepts are more abstract.

It is interesting to note that while the $\ell_{\textsc{match}}$ used in ViLBERT w/ DG incorporates learning representations to align images at both specific and abstract levels, such learning benefits all levels. The improvement of retrieving at abstract levels does not sacrifice the retrieval at specific levels.

% !TEX root = main_emnlp2020.tex
\section{Conclusion}
\label{sConclusion}
Image and text aligned data is rich in semantic correspondence. Besides treating text annotations as ``categorical'' labels, in this paper, we show that we can make full use of those labels. Concretely, denotation graphs (DGs) encode structural relations that can be automatically extracted from those texts with linguistic analysis tools.  We proposed several ways to incorporate DGs into learning representation and validated the proposed approach on several tasks. We plan to investigate other automatic tools in curating more accurate denotation graphs with a complex composition of fine-grained concepts for future directions.

\section*{Acknowledgement}
{We appreciate the feedback from the reviewers. This work is partially supported by NSF Awards IIS-1513966/ 1632803/1833137, CCF-1139148, DARPA Award\#: FA8750-18-2-0117, FA8750-19-1-0504
, DARPA-D3M - Award UCB-00009528, Google Research Awards, gifts from Facebook and Netflix, and ARO\# W911NF-12-1-0241 and W911NF-15-1-0484. We particularly thank Haoshuo Huang for help in improving the efficiency of DG generation.}

\bibliographystyle{acl_natbib}
\bibliography{main}

\clearpage
\appendix

\section*{Appendix}

In the Appendix, we provide details omitted from the main text due to the limited space, including:
\begin{itemize}[topsep=1pt,parsep=0pt,partopsep=4pt,leftmargin=*,itemsep=2pt]
    \item \S~\ref{supp:sec:impl} describes complete implementation details (cf. \S~3 and \S~5.1 of the main text).
    \item \S~\ref{supp:sec:exp} provides complete experimental results (cf. \S~5.2 of the main text).
    \item \S~\ref{supp:sec:vis}  visualizes the model's predictions on denotation graphs.
\end{itemize}

\section{Implementation Details}
\label{supp:sec:impl}
% !TEX root = supp_emnlp2020.tex

\subsection{Constructing Denotation Graphs}

We summarize the procedures used to extract DG from \vl datasets. For details, please refer to~\cite{young2014image}. We used the publicly available tool\footnote{\url{https://github.com/aylai/DenotationGraph}}. The analysis consists of several steps: (1) spell-checking; (2) tokenize the sentences into words; (3) tag the words with Part-of-Speech labels and chunk works into phrases; (4) abstract semantics by using the WordNet~\cite{miller1995wordnet} to construct a hypernym lexicon table to replace the nouns with more generic terms; (5) apply 6 types of templated rules to create fine-to-coarse (\ie, specific to generic) semantic concepts and connect the concepts with edges.

We set 3 as the maximum levels (counting from the sentence level) to extract abstract semantic concepts. This is due to the computation budget we can afford, as the final graphs can be huge in both the number of nodes and the edges. Specifically, without the maximum level constraint, we have 2.83$M$ concept nodes in total for Flickr dataset. If the training is run on all these nodes, we will consume 19 times more iterations than training on the original dataset, which has 145K sentences~\cite{young2014image}. As a result, much more time would be required for every experiment. With the 3 layers of DG from the leaf concepts, we have in 597K nodes. In this case, the training time would be cut down to 4.1 times of the original dataset.

Nonetheless, we experimented with more than 3 levels to train ViLBERT + \DGFlickr with 5 and 7 maximum levels, respectively. The training hyper-parameters remain the same as ViLBERT + \DGFlickr with 3 maximum layers. The aim is to check how much gain we could get from the additional annotations. We report the results in Table~\ref{tab:flickr_dgraph_layer}. It shows that actually, the model trained with 3 levels of DG achieves the best performance. This might be because those high-level layers of DG (counting from the sentences) contain very abstract text concepts, such as ``entity'' and ``physical object'', which is non-informative in learning the visual grounding.

\begin{table}[t]
    \centering
    \tabcolsep 4.5pt
    \caption{
    	Text-based Image Retrieval Performance of ViLBERT trained with different number of DG levels
    }
     \begin{tabular}{@{\;\;}l c c c c@{\;\;}}
        \toprule
        {\bf \# of DG levels} & R@1 & R@5 & R@10 & RSUM \\
        \midrule
        3 levels & 65.9 & 91.4 & 95.5 & 252.7 \\
        5 levels & 62.5 & 86.4 & 92.3 & 241.2 \\
        7 levels & 62.8 & 86.3 & 91.6 & 240.7 \\
        \bottomrule
    \end{tabular}
    \label{tab:flickr_dgraph_layer}
\end{table}

Once the graph is constructed, we attach the images to the proper nodes by set-union images of each node's children, starting from the sentence-level node.

\subsection{Model architectures of ViLBERT and UNITER}

\begin{figure*}[t]
    \centering
    \includegraphics[width=1.0\linewidth]{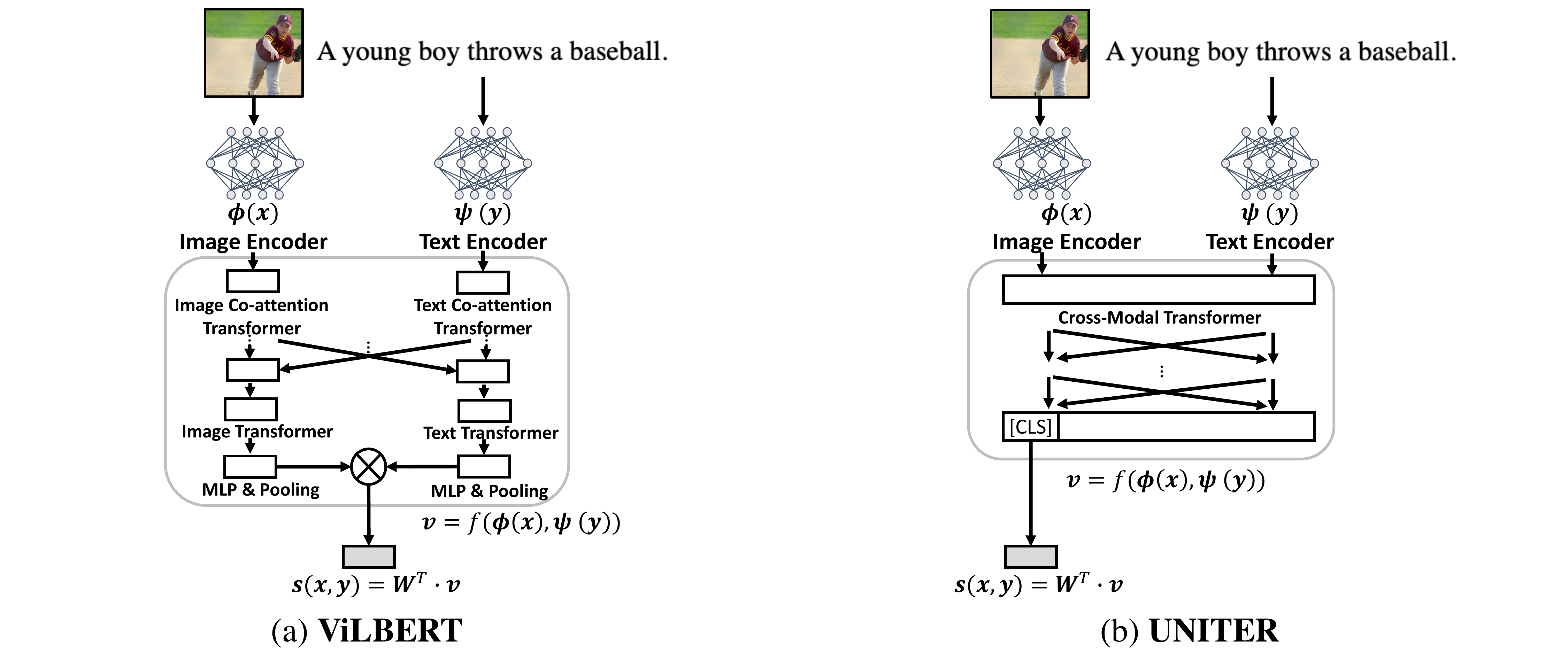}

    \caption{Architecture of (a) ViLBERT, (b) UNITER. The $\bigotimes$ means element-wise product. The [CLS] represents the embedding of [CLS] token in the last UNITER layer.
    }
    \label{fig:arch}
\end{figure*}

A comparison of these models is schematically illustrate in Fig.~\ref{fig:arch}.

\begin{itemize}
\item ViLBERT. It has 6 basic Transformer layers for text and 8 layers for image. For all the Transformer layers on the text side, we use 12 attention heads and 256 feature dimensions, then linearly project down to 1024 feature dimensions. For all the Transformers on the image side, we use 8 attention heads and 128 feature dimensions, then combine into 1024 feature dimensions too.

\item UNITER. All the Transformer layers have 12 heads and 256 feature dimensions.
\end{itemize}

The major difference between UNITER and ViLBERT is how attentions are used. In ViLBERT, one modality is used as a query, and the other is used as value and key. In UNITER, however, both are used as query, key, and value. Additionally, UNITER is similar to another model Unicoder-VL~\cite{Li2019UnicoderVLAU}. However, the latter has not provided publicly available code for experimenting.

For ViLBERT model, each text and image co-attention Transformer layer contains 8 attention heads with 1024 dimensions in total. The text Transformer layer contains 12 attention heads with 3072 hidden dimensions in total. In contrast, the image Transformer layer has 8 attention heads with 1024 hidden dimensions in total.
For UNITER model, each cross-attention Transformer layer contains 12 heads with 3072 hidden dimensions in total.

ViLBERT model contains 121 million parameters, while UNITER contains 111 million parameters.

\subsection{Training Details}

All models are optimized with the Adam optimizer~\cite{kingma2014adam}. The learning rate is initialized as $4e^{-5}$. Following ViLBERT~\cite{Lu2019ViLBERTPT}, a warm-up training session is employed, during which we linearly increase the learning rate from $0$ to $4e^{-5}$ in the first $1.5\%$ part of the training epochs. The learning rate is dropped to $4e^{-6}$ and $4e^{-7}$ at the $10$th and the $15$th epochs, respectively. For ViLBERT (Reduced), we randomly initialized the model parameters in the image stream. The text stream is initialized from the first 3 layers of the pre-trained BERT model, and its co-attention Transformer layers are randomly initialized. For ViLBERT (Full) and UNITER~\cite{chen2019uniter}, we load the model's weights pre-trained on the Conceptual Caption dataset to initialize them.

Training ViLBERT (Full) + DG with a minibatch size of 64 takes 2 to 3 days on an 8 TitanXp GPU server, or 1 day on TPU v2 cloud. The GPU server is equipped with Intel Xeon Gold 6154 CPU and 256G RAM.

\subsection{Text Pre-processing}
We follow BERT~\cite{Devlin2019BERT} that uses WordPiece~\cite{wu2016google} tokenizer to tokenize the texts. For ViLBERT (Reduced) and ViLBERT (Full), we use the uncased tokenizer with a vocabulary size of 30,522. For UNITER, we use the cased tokenizer with a vocabulary size of 28,996. After tokenization, the tokens are transformed to 768 dimension features by a word embedding initialized from BERT pre-trained model. The 768-dimensional position features are included in the input to represent the position of each token.

\subsection{Visual Pre-processing}
For both ViLBERT and UNITER, we use the image patch features generated by the bottom-up attention features, as suggested by the original papers~\cite{anderson2017updown}.
The image patch features contain up to $100$ image patches with their dimensions to be 2048.
Besides this, a positional feature is used to represent the spatial location of bounding boxes for both ViLBERT and UNITER. Specifically, ViLBERT uses 5-dimensional position feature that encodes the normalized coordinates of the upper-left and lower-right corner for the bounding boxes, as well as one additional dimension encoding the normalized patch size. UNITER uses two additional spatial features that encode the normalized width and height of the object bounding box.

\section{Full Experimental Results}
\label{supp:sec:exp}
% !TEX root = supp_emnlp2020.tex

In this section, we include additional experimental results referred to by the main text. Specifically, we include results from a variety of models (\eg, ViLBERT, ViLBERT + DG, UNITER, and UNITER + DG) on \COCO dataset 5K test split~\cite{karpathy2015deep} in \S~\ref{sec:full_result}. Then we provide a comprehensive ablation study on the impact of $\lambda_1$ and $\lambda_2$ of Eq. 7 in the main text in \S~\ref{sec:abs_hyper}.

\subsection{Complete Results on \COCO Dataset}
\label{sec:full_result}
We report the full results on \COCO dataset (1K test split and 5K test split) in Table \ref{tab:coco_5k} and Table~\ref{tab:coco_5k_sentence_retrieval}. Additionally, we contrast to other existing approaches on these tasks.
It could be seen that ViLBERT + DG and UNITER + DG improves the performance over the counterparts without DG by a significant margin on both \COCO 1K and 5K test split -- the only exception is that on the task of image-based text retrieval, UNITER performs better than UNITER+DG.

These results support our claim that training with DG helps the model to learn better visual and linguistic features. Although ViLBERT and UNITER have different architectures, training with DG could improve the performance consistently. 
\begin{table*}[t]
    \centering
    \tabcolsep 1pt
    \caption{
        Results on Cross-Modal Retrieval on \COCO dataset 1K test split (Higher is better)
    }
    \begin{tabular}{cc}
        Text-based Image Retrieval & Image-based Text Retrieval  \\
        \begin{minipage}[t]{0.5\textwidth}
        \resizebox{\linewidth}{!}
        {
            \tabcolsep 2pt
            \begin{tabular}{@{\;}l@{\;}cccc @{\;}}
                \addlinespace
                \toprule
                Method & \textsc{r@1} & \textsc{r@5} & \textsc{r@10} & \textsc{rsum} \\
                 \midrule
                \multicolumn{5}{@{\;}l}{\bf \scriptsize{Models ran or implemented by us}} \\
        ViLBERT & 62.3 & 89.5 & 95.0 & 246.8 \\
        ViLBERT + DG & 65.9 & 91.4 & 95.5 & 252.7 \\
        UNITER  &  60.7 & 88.0 & 93.8 & 242.5 \\
        UNITER + DG &  62.7 & 88.8 & 94.4 & 245.9 \\
                \midrule
                \multicolumn{5}{@{\;}l}{\bf \scriptsize{Known results from literature}} \\
                VSE++\cite{faghri2018vse++}  & 52.0 & 84.3 & 92.0 & 228.3 \\
                SCO\cite{huang2018SCO}       & 56.7 & 87.5 & 94.8 & 239.0 \\
                SCAN\cite{lee2018scan}       & 58.8 & 88.4 & 94.8 & 242.0 \\
                VSRN\cite{Li2019VSRN}        & 62.8 & 89.7 & 95.1 & 247.6 \\
                \bottomrule
            \end{tabular}
        }
        \end{minipage} &
        \begin{minipage}[t]{0.5\textwidth}
        \resizebox{\linewidth}{!}
        {
            \tabcolsep 2pt
            \begin{tabular}{@{\;}l@{\;}cccc @{\;}}
                \addlinespace
                \toprule
                Method & \textsc{r@1} & \textsc{r@5} & \textsc{r@10} & \textsc{rsum} \\
                 \midrule
                \multicolumn{5}{@{\;}l}{\bf \scriptsize{Models ran or implemented by us}} \\
                ViLBERT      &  77.0 & 94.1 & 97.2 & 268.3 \\
                ViLBERT + DG & 79.0 & 96.2 & 98.6 & 273.8\\
                UNITER       & 74.4 & 93.9 & 97.1 & 265.4 \\
                UNITER + DG  & 77.7 & 95.0 & 97.5 & 270.2 \\
                \midrule
                \multicolumn{5}{@{\;}l}{\bf \scriptsize{Known results from literature}} \\
                VSE++\cite{faghri2018vse++}  &  64.6 & 90.0 & 95.7 & 250.3 \\
                SCO\cite{huang2018SCO}       &  69.9 & 92.9 & 97.5 & 260.3 \\
                SCAN\cite{lee2018scan}       &  72.7 & 94.8 & 98.4 & 265.9 \\
                VSRN\cite{Li2019VSRN}        &  76.2 & 94.8 & 98.2 & 269.2 \\
                \bottomrule
            \end{tabular}
        }
        \end{minipage}
    \end{tabular}
    \label{tab:coco_5k}
\end{table*}

\begin{table*}[t]
    \centering
    \tabcolsep 1pt
    \caption{
        Results on Cross-Modal Retrieval on \COCO dataset 5K test split (Higher is better)
    }
    \begin{tabular}{cc}
        Text-based Image Retrieval & Image-based Text Retrieval  \\
        \begin{minipage}[t]{0.5\textwidth}
        \resizebox{\linewidth}{!}
        {
            \tabcolsep 2pt
            \begin{tabular}{@{\;}l@{\;}cccc @{\;}}
                \addlinespace
                \toprule
                Method & \textsc{r@1} & \textsc{r@5} & \textsc{r@10} & \textsc{rsum} \\
                 \midrule
                \multicolumn{5}{@{\;}l}{\bf \scriptsize{Models ran or implemented by us}} \\
        ViLBERT & 38.6 & 68.2 & 79.0 & 185.7 \\
        ViLBERT + DG & 41.8 & 71.5 & 81.5 & 194.8 \\
        UNITER & 37.8 & 67.3 & 78.0 & 183.1 \\
        UNITER + DG &  39.1 & 68.0 & 78.3 & 185.4 \\
                \midrule
                \multicolumn{5}{@{\;}l}{\bf \scriptsize{Known results from literature}} \\
                VSE++\cite{faghri2018vse++}  & 30.3 & 59.4 & 72.4 & 162.1 \\
                SCO\cite{huang2018SCO}       & 33.1 & 62.9 & 75.5 & 171.5 \\
                SCAN\cite{lee2018scan}       & 38.6 & 69.3 & 80.4 & 188.3 \\
                VSRN\cite{Li2019VSRN}        & 40.5 & 70.6 & 81.1 & 192.2 \\
                UNITER\cite{chen2019uniter}$^\dagger$ & 48.4 & 76.7 & 85.9 & 211.0 \\
                \bottomrule
            \end{tabular}
        } 
        \end{minipage} &
        \begin{minipage}[t]{0.5\textwidth}
        \resizebox{\linewidth}{!}
        {
            \tabcolsep 2pt
            \begin{tabular}{@{\;}l@{\;}cccc @{\;}}
                \addlinespace
                \toprule
                Method & \textsc{r@1} & \textsc{r@5} & \textsc{r@10} & \textsc{rsum} \\ \midrule
                \multicolumn{5}{@{\;}l}{\bf \scriptsize{Models ran or implemented by us}} \\
                ViLBERT      & 53.5 & 79.7 & 87.9 & 221.1\\
                ViLBERT + DG & 57.5 & 84.0 & 90.1 & 232.2 \\
                UNITER       & 52.8 & 79.7 & 87.8 & 220.3 \\
                UNITER + DG  & 51.4 & 78.7 & 87.0 & 217.1 \\
                \midrule
                \multicolumn{5}{@{\;}l}{\bf \scriptsize{Known results from literature}} \\
                VSE++\cite{faghri2018vse++}          &  41.3 & 71.1 & 81.2 & 193.6 \\
                SCO\cite{huang2018SCO}               &  42.8 & 72.3 & 83.0 & 198.1 \\
                SCAN\cite{lee2018scan}               &  50.4 & 82.2 & 90.0 & 222.6 \\
                VSRN\cite{Li2019VSRN}                &  53.0 & 81.1 & 89.4 & 223.5 \\
                UNITER \cite{chen2019uniter}$^\dagger$ &  63.3 & 87.0 & 93.1 & 243.4 \\
                \bottomrule
            \end{tabular}
        }
        \end{minipage}
    \end{tabular}
    \label{tab:coco_5k_sentence_retrieval}
     \\
    {\small \begin{flushleft}
    $^\dagger$: The UNITER\cite{chen2019uniter} model performs an additional online hard-negative mining (which we did not) during the training of image-text matching to improve their results, which is computationally very costly.
    \end{flushleft}}

\end{table*}

\subsection{Complete Results on \Flickr Dataset}
\begin{table*}[t]
    \centering
    % \resizebox{\linewidth}{!}
    \tabcolsep 2pt
    \caption{Results on Text-based Image Retrieval on \Flickr test split (Higher is better)}
    \begin{tabular}{@{\;}l@{\;}cccc @{\;}}
        \addlinespace
        \toprule
        Method &\textsc{r@1} & \textsc{r@5} & \textsc{r@10} & \textsc{rsum} \\
        \midrule
        \multicolumn{5}{@{\;}l}{\bf \scriptsize{Models ran or implemented by us}} \\
        ViLBERT & 59.1 & 85.7 & 92.0 & 236.7  \\
        ViLBERT + DG & 63.8 & 87.3 & 92.2 & 243.3 \\
        UNITER  & 62.9 & 87.2 & 92.7 & 242.8 \\
        UNITER + DG & 66.4 & 88.2 & 92.2 & 246.8 \\
       	\midrule
		\multicolumn{5}{@{\;}l}{\bf \scriptsize{Known results from literature}} \\
        VSE++\cite{faghri2018vse++} & 39.6 & 70.1 & 79.5 & 189.2 \\
        SCO\cite{huang2018SCO} & 41.1 & 70.5 & 80.1 & 191.7 \\
        SCAN\cite{lee2018scan} & 48.6 & 77.7 & 85.2 & 211.5 \\
        VSRN\cite{Li2019VSRN} & 54.7 & 81.8 & 88.2 & 224.7 \\
        ViLBERT\cite{Lu2019ViLBERTPT} & 58.2 & 84.9 & 91.5 & 234.6 \\
        UNITER\cite{chen2019uniter} & 71.5 & 91.2 & 95.2 & 257.9 \\
        \bottomrule
	\end{tabular}
    \label{tab:flickr}
\end{table*}

We contrast to other existing approaches in Table~\ref{tab:flickr} on the task of text-based image retrieval on the \Flickr dataset.

\subsection{Ablation Study on $\lambda_1$ and $\lambda_2$}
\label{sec:abs_hyper}

We conduct an ablation study on the impact of the two hyper-parameters $\lambda_1$ and $\lambda_2$ in Eq. 7 of the main text. We conduct the study with two ViLBERT variants: ViLBERT Reduced and ViLBERT. The results are reported in Table \ref{tab:vilbert_reduced} and Table \ref{tab:vilbert}. As we have two hyper-parameters $\lambda_1$ and $\lambda_2$, we analyze their impacts on the final results by fixing one $\lambda$ to be $1$. Fixing the $\lambda_2 =1$ and changing $\lambda_1$, we observe that ViLBERT prefers larger $\lambda_1$, while ViLBERT Reduced achieves slightly worse performance when $\lambda_1$ is smaller or larger. Fixing the $\lambda_1 =1$ and changing $\lambda_2$, we observe that performance of both architectures slightly reduced when $\lambda_2 = 0.5$ and $\lambda_2 = 2$.

\begin{table*}[t]
    \centering
    \tabcolsep 1pt
    \caption{
        Ablation studies on the impact of $\lambda_1$ and $\lambda_2$ of ViLBERT Reduced on Text-based Image Retrieval on \Flickr dataset (Higher is better)
    }
    \begin{tabular}{cc}
        (a) Ablating $\lambda_1$ & (b) Ablating $\lambda_2$ \\
        \begin{minipage}[t]{0.5\textwidth}
        \centering
        {
            \tabcolsep 4pt
            \begin{tabular}{@{\;}ll@{\quad}ccc@{\quad}c @{\;}}
                \addlinespace
                \toprule
                $\lambda_1$ & $\lambda_2$ & \textsc{r@1} & \textsc{r@5} & \textsc{r@10} & \textsc{rsum} \\
                 \midrule
                $0.5$ & $1.0$ &  57.7 & 83.1 & 88.5 & 229.2 \\
                $1.0$ & $1.0$ & 58.7 & 83.3 & 89.3 & 231.2 \\
                $2$ & $1.0$ &  56.5 & 82.6 & 88.6 & 227.7 \\
                \bottomrule
            \end{tabular}
        }
        \end{minipage} &
        \begin{minipage}[t]{0.5\textwidth}
        \centering
        {
            \centering
            \tabcolsep 4pt
            \begin{tabular}{@{\;}ll@{\quad}ccc@{\quad}c @{\;}}
                \addlinespace
                \toprule
                $\lambda_1$ & $\lambda_2$ & \textsc{r@1} & \textsc{r@5} & \textsc{r@10} & \textsc{rsum} \\
                 \midrule
                $1.0$ & $0.5$ & 56.3 & 81.7 & 87.2 & 225.2 \\
                $1.0$ & $1.0$ & 58.7 & 83.3 & 89.3 & 231.2 \\
                $1.0$ & $2$ & 58.5 & 82.3 & 88.0 & 228.9 \\
                \bottomrule
            \end{tabular}
        }
        \end{minipage}
    \end{tabular}
    \label{tab:vilbert_reduced}
\end{table*}

\begin{table*}[t]
    \centering
    \tabcolsep 1pt
    \caption{
        Ablation studies on the impact of $\lambda_1$ and $\lambda_2$ of ViLBERT on Text-based Image Retrieval on \Flickr dataset (Higher is better)
    }
    \begin{tabular}{cc}
        (a) Ablating $\lambda_1$ & (b) Ablating $\lambda_2$ \\
        \begin{minipage}[t]{0.5\textwidth}
        \centering
        {
            \tabcolsep 4pt
            \begin{tabular}{@{\;}ll@{\quad}ccc@{\quad}c @{\;}}
                \addlinespace
                \toprule
                $\lambda_1$ & $\lambda_2$ & \textsc{r@1} & \textsc{r@5} & \textsc{r@10} & \textsc{rsum} \\
                 \midrule
                $0.5$ & $1.0$ & 63.1 & 86.7 & 91.7 & 241.4 \\
                $1.0$ & $1.0$ & 63.8 & 87.3 & 92.2 & 243.3 \\
                $2$ & $1.0$ & 64.1 & 87.6 & 92.5 & 244.2 \\
                \bottomrule
            \end{tabular}
        }
        \end{minipage} &
        \begin{minipage}[t]{0.5\textwidth}
        \centering
        {
            \tabcolsep 4pt
            \begin{tabular}{@{\;}ll@{\quad}ccc@{\quad}c @{\;}}
                \addlinespace
                \toprule
                $\lambda_1$ & $\lambda_2$ & \textsc{r@1} & \textsc{r@5} & \textsc{r@10} & \textsc{rsum} \\
                 \midrule
                $1.0$ & $0.5$ & 63.7 & 87.0 & 92.4 & 243.2 \\
                $1.0$ & $1.0$ & 63.8 & 87.3 & 92.2 & 243.3 \\
                $1.0$ & $2$ & 63.1 & 86.6 & 91.9 & 241.6 \\
                \bottomrule
            \end{tabular}
        }
        \end{minipage}
    \end{tabular}
    \label{tab:vilbert}
\end{table*}

\subsection{Full Results on Zero/Few-Shot and Transfer Learning}

\mypara{Implementation Details for Zero-shot Referring Expression}
Specifically, the learned ViLBERT and ViLBERT w/DG models are used first to produce a base matching score $\vct{s}_{\textsc{base}}$  between the expression to be referred and the whole image. We then compute the matching score $\vct{s}_{\textsc{masked}}$ between the expression and the image with each region feature being replaced by a random feature in turn. As the masked image region might be a noisy region, $\vct{s}_{\textsc{masked}}$ might be larger than $\vct{s}_{\textsc{base}}$. Therefore, the model's prediction of which region the expression refers to is the masked region which causes the largest score in $\vct{s}_{\textsc{region}}$, where
\[
\vct{s}_{\textsc{region}} = (\vct{s}_{\textsc{base}} - \vct{s}_{\textsc{masked}}) \cdot \mathbb{I}[\vct{s}_{\textsc{masked}} > \vct{s}_{\textsc{base}}].
\]
Here $\mathbb{I}[\cdot]$ is an indicator function. Table~\ref{tZeroFewShot} shows that ViLBERT + \DGCOCO outperforms  ViLBERT on this task.

\mypara{Transfer Learning Results} Table~\ref{tab:full_transfer} reports the full set of evaluation metrics on transferring across datasets. Training with DG improves training without DG noticeably.

\begin{table*}[t]
    \centering
    \tabcolsep 4pt
    \caption{Transferrability of the learned representations}
    \begin{tabular}{@{\;}l@{\quad}cccc @{\quad} cccc@{\;}}
        \addlinespace
        \toprule
        \textsc{source}$\rightarrow$\textsc{target} & \multicolumn{4}{c}{\Flickr$\rightarrow$\COCO} & \multicolumn{4}{c}{\COCO$\rightarrow$\Flickr} \\
         Model & \textsc{r@1} & \textsc{r@5} & \textsc{r@10} & \textsc{rsum} & \textsc{r@1} & \textsc{r@5} & \textsc{r@10} & \textsc{rsum} \\
        \midrule
        ViLBERT &  43.5 & 72.5 & 83.4 & 199.4 & 49.0 & 76.0 & 83.9 & 209.0  \\
        ViLBERT + \textsc{source} DG & 44.9 & 72.7 & 83.0 & 200.5 & 52.8 & 79.2 & 86.2 & 218.2 \\
        \bottomrule
    \end{tabular}
    \label{tab:full_transfer}
\end{table*}

\section{Visualization of Model's Predictions on Denotation Graphs}
\label{supp:sec:vis}
% !TEX root = supp_emnlp2020.tex

\begin{figure*}[t]
	\centering
\includegraphics[width=1\textwidth]{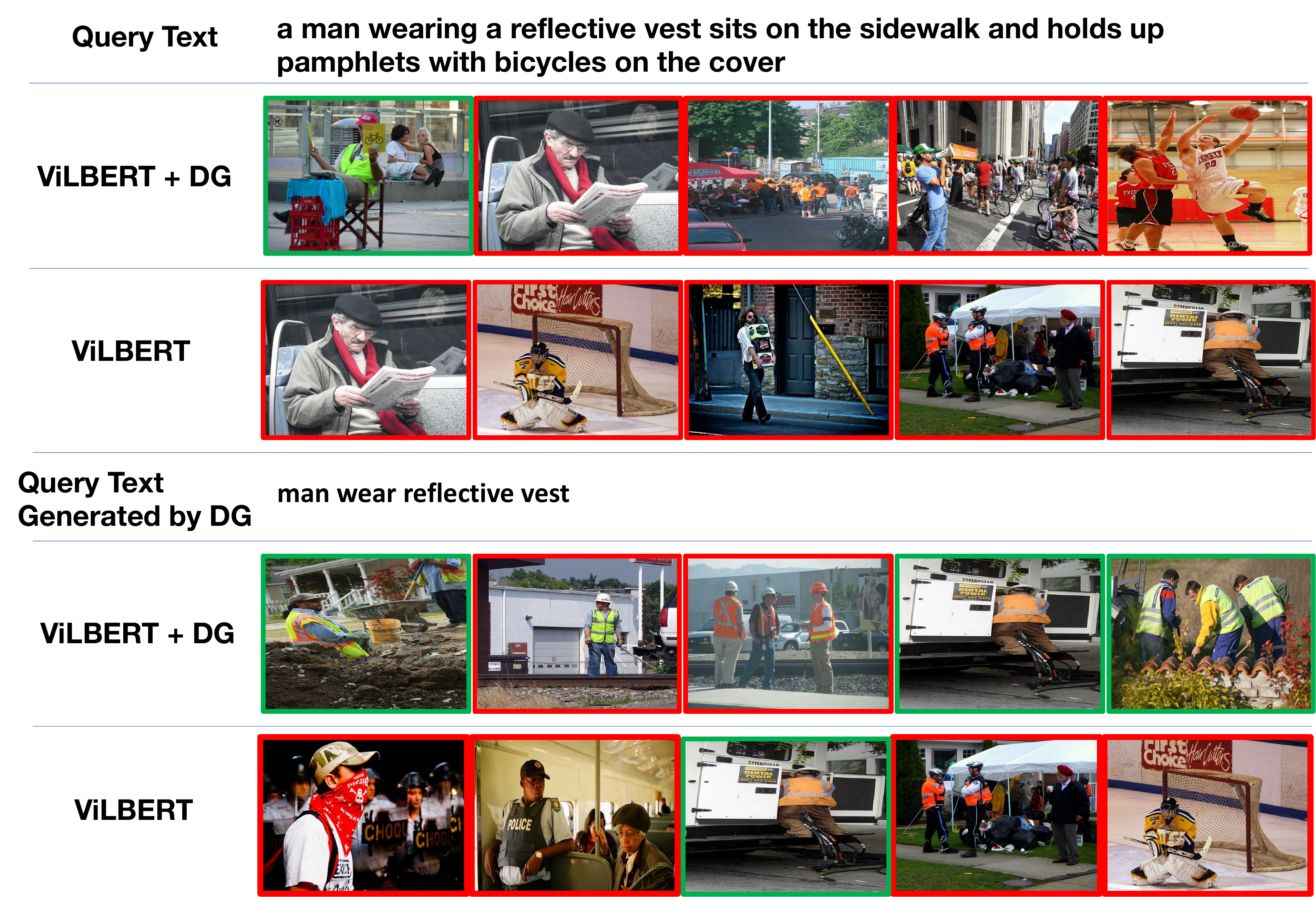}
	\caption{{\small \textbf{\Flickr Denotation Graph: Given Text and Retrieve Image.} Qualitative example of ViLBERT + DG successfully retrieves the text matched images. We mark the correct sample in \textbf{ \color{green} green} and incorrect one in \textbf{ \color{red} red}.} }
	\label{fig:succ_case}
\end{figure*}

\begin{figure*}[t]
	\centering
\includegraphics[width=1\textwidth]{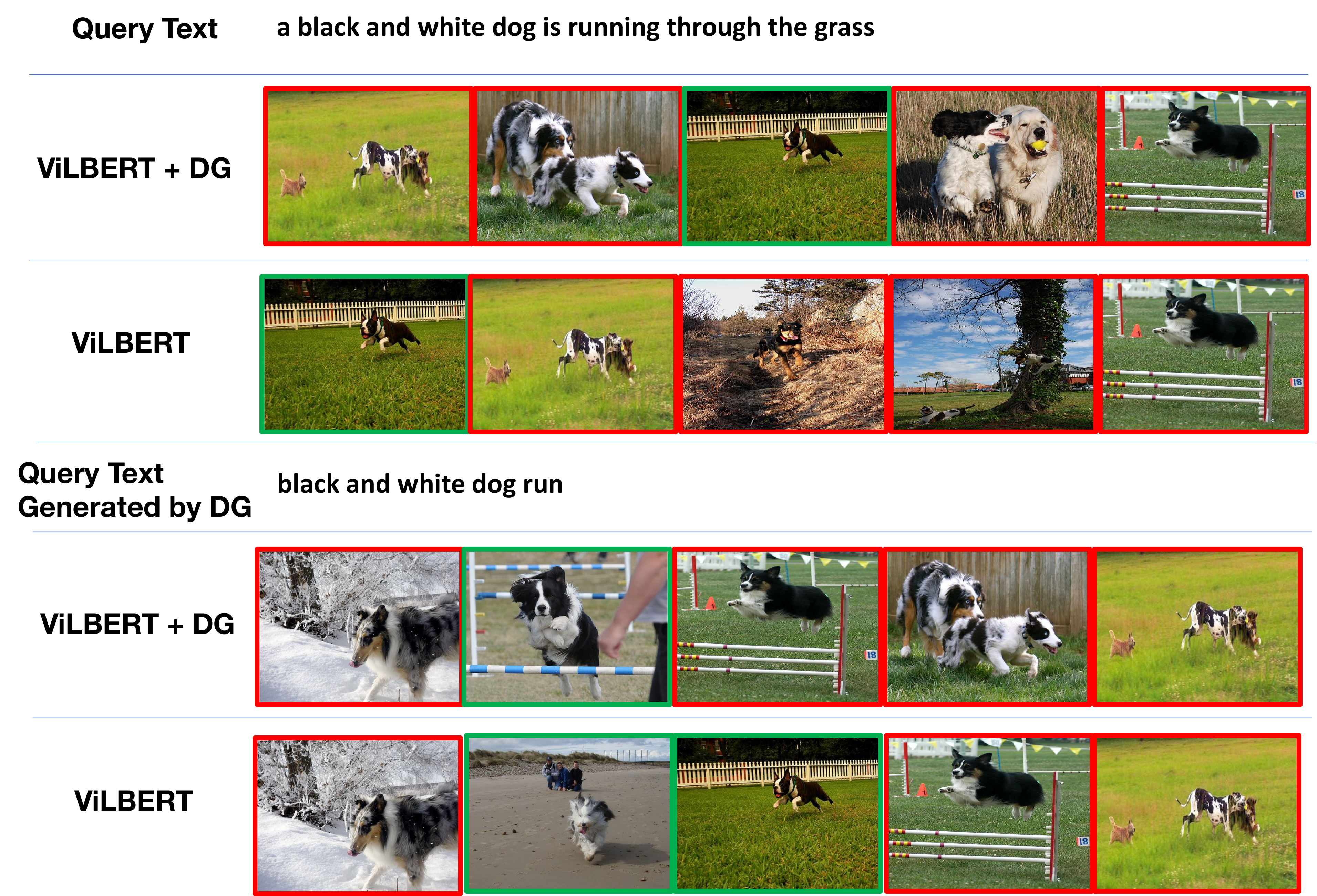}
	\caption{{\small \textbf{\Flickr Denotation Graph: Given Text and Retrieve Image.} Qualitative example of ViLBERT + DG fails to retrieve the text matched images. We mark the correct sample in \textbf{ \color{green} green} and incorrect one in \textbf{ \color{red} red}.} }
	\label{fig:fail_case}
\end{figure*}

We show several qualitative examples of both success and failure cases of ViLBERT + DG, when retrieving the text matched images,  in Fig.~\ref{fig:succ_case} and Fig.~\ref{fig:fail_case}. The image and text correspondence is generated by the Denotation Graph, which are derived from the caption and image alignment. We observe that in the Fig.\ref{fig:succ_case}, the ViLBERT + DG successfully recognizes the images that are aligned with the text: ``man wear reflective vest'', while the ViLBERT fails to retrieve the matched image. In the failure case in Fig. \ref{fig:fail_case}, although ViLBERT + DG fails to retrieve the images that are exactly matched to the text, it still retrieves very relevant images given the query.

\end{document}